\documentclass[runningheads]{llncs}

\usepackage{eccv}

\usepackage{eccvabbrv}

\usepackage{graphicx}
\usepackage{booktabs}

\usepackage{tabularx}
\usepackage{multirow}
\usepackage{array} 

\graphicspath{{./figures}}
\DeclareGraphicsExtensions{.pdf,.jpeg,.jpg,.png}

\newcolumntype{V}{>{\raggedright\arraybackslash}X}
\newcolumntype{Y}{>{\centering\arraybackslash}X}
\newcolumntype{Z}{>{\raggedleft\arraybackslash}X}

\newcolumntype{v}{>{\hsize=.5\hsize}V}
\newcolumntype{y}{>{\hsize=.5\hsize}Y}
\newcolumntype{z}{>{\hsize=.5\hsize}Z}

\usepackage[accsupp]{axessibility}  

\usepackage{hyperref}

\usepackage{orcidlink}

\begin{document}

\title{Down-Sampling Inter-Layer Adapter for Parameter and Computation Efficient Ultra-Fine-Grained Image Recognition} 

\vspace{-3cm}

\titlerunning{Down-Sampling Inter-Layer Adapter}

\author{Edwin Arkel Rios\inst{1} \and
Femiloye Oyerinde\inst{2} \and
Min-Chun Hu\inst{3} \and
Bo-Cheng Lai\inst{1}}

\authorrunning{E.A.Rios et al.}

\institute{National Yang Ming Chiao Tung University, Taiwan \and
Cohere For AI Community \and
National Tsing Hua University, Taiwan
}

\maketitle

\vspace{-0.25cm}

\begin{abstract}
    Ultra-fine-grained image recognition (UFGIR) categorizes objects with extremely small differences between classes, such as distinguishing between cultivars within the same species, as opposed to species-level classification in fine-grained image recognition (FGIR). The difficulty of this task is exacerbated due to the scarcity of samples per category. To tackle these challenges we introduce a novel approach employing down-sampling inter-layer adapters in a parameter-efficient setting, where the backbone parameters are frozen and we only fine-tune a small set of additional modules. By integrating dual-branch down-sampling, we significantly reduce the number of parameters and floating-point operations (FLOPs) required, making our method highly efficient. Comprehensive experiments on ten datasets demonstrate that our approach obtains outstanding accuracy-cost performance, highlighting its potential for practical applications in resource-constrained environments. In particular, our method increases the average accuracy by at least 6.8\% compared to other methods in the parameter-efficient setting while requiring at least 123x less trainable parameters compared to current state-of-the-art UFGIR methods and reducing the FLOPs by 30\% in average compared to other methods.
\end{abstract}

\section{Introduction}
\label{sec_intro}

Ultra-fine-grained image recognition (UFGIR) categorizes sub-categories within a macro-category. While conventional FGIR \cite{wei_fine-grained_2021} classifies objects usually up to species-level granularity, UFGIR may categorize classes at a finer level, such as cultivars of a plant. It has practical application in various fields such as agriculture \cite{larese_multiscale_2014, mohanty_using_2016}, medical \cite{park_fine-grained_2023}, and industrial \cite{lehr_automated_2020}. It is a challenging task due to small inter-class differences, large intra-class differences, and low data availability due to the difficulty behind labeling even for human experts \cite{yu_benchmark_2021}.

To address this most methods \cite{yu_patchy_2020,yu_maskcov_2021,wang_feature_2021,yu_spare_2022,yu_cle-vit_2023,yu_mix-vit_2023} utilize coarse image-recognition backbones equipped with additional modules \cite{he_transfg_2022,wang_feature_2021} or losses \cite{yu_maskcov_2021,fang_learning_2024} to focus and make better use of discriminative features that encapsulate subtle differences between fine-grained classes. Specifically, recent works employ Vision Transformers (ViT) \cite{dosovitskiy_image_2020} since their use of the self-attention  mechanism \cite{vaswani_attention_2017}, with its global receptive field, allows models to effectively extract and aggregate fine-grained features \cite{hu_rams-trans_2021,sun_sim-trans_2022,rios_anime_2022}.

However, due to the growing size of state-of-the-art (SotA) ViT backbones, researchers have explored the design of parameter-efficient transfer learning (PETL) methods \cite{houlsby_parameter-efficient_2019,avidan_improving_2022,jie_convolutional_2022,tu_visual_2023}. Instead of fine-tuning the entire backbone, most of the parameters are frozen, and only specific components are fine-tuned. This allows for reuse of most of the parameters, drastically reducing storage requirements, specially when deploying models across multiple tasks.

PETL methods have shown performance that can match or even surpass specialized FGIR models with full fine-tuning in generic FGIR tasks \cite{avidan_improving_2022}. However, we observe that in UFGIR tasks PETL methods still lag behind specialized FGIR methods, either in the traditional fine-tune setting or in a novel setting that we coin as parameter-efficient FGIR (PEFGIR), where we only fine-tune the FGIR modules while most of the backbone is frozen. Based on analysis of the ViT features, we observe on \cref{figure_cka_attention} that frozen ViTs suffer from attention collapse \cite{zhou_deepvit_2021,wang_anti-oversmoothing_2022,chen_principle_2022}, a phenomenona where the attention scores across layers become increasingly similar, hindering the feature extraction process.

\begin{figure}[!htb]
    \vspace{-0.5cm}
    \begin{center}
        \includegraphics[width=0.6\linewidth]{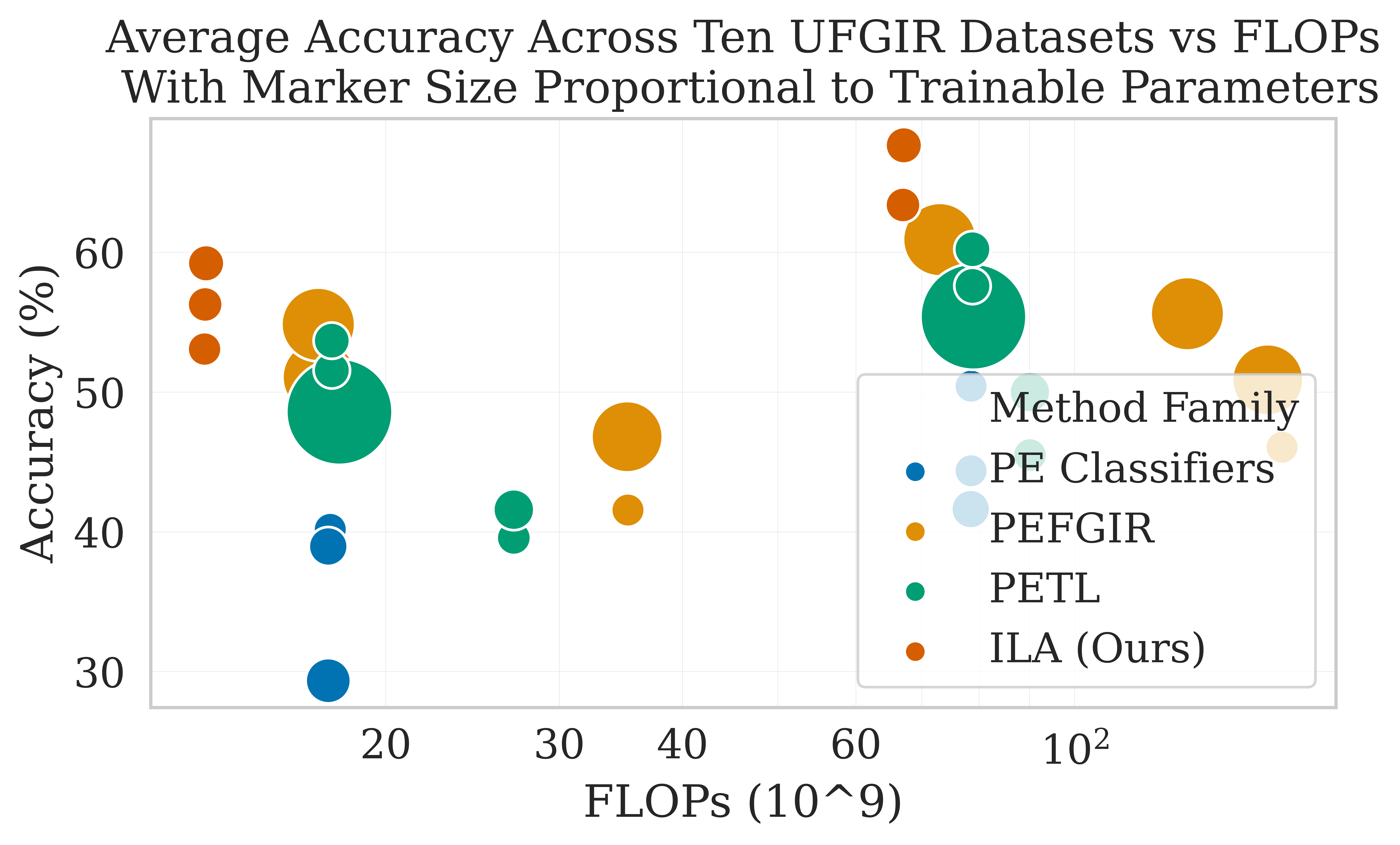}
    \end{center}
    \vspace{-0.5cm}
    \caption{Average top-1 accuracy (\%) across all evaluated datasets vs number of floating-point operations (FLOPs) for different method families, including methods that only fine-tune the classification head, fine-grained image recognition (FGIR) methods in parameter-efficient setting (PEFGIR, only fine-tune the fine-grained discrimination modules) and parameter-efficient transfer learning (PETL) methods. The size of the markers is proportional to the percentage of trainable parameters for each method.}
    \vspace{-0.5cm}
    \label{figure_accuracy_vs_flops_vs_params}
\end{figure}

To address this, we propose Intermediate Layer Adapter (ILA), a novel, parameter and computationally efficient method that employs dual spatial down-sampling branches as an adapter \cite{houlsby_parameter-efficient_2019,jie_convolutional_2022} inserted between transformer layers to aggregate spatial features while preserving fine-grained details. Our proposed approach results in much more diverse attention scores across layers and across a variety of benchmarks we demonstrate its outstanding performance compared to SotA FGIR and PETL methods in terms of accuracy and computational cost, as seen in \cref{figure_accuracy_vs_flops_vs_params}.

Our contributions are as follows:
\begin{enumerate}
    \vspace{-0.25cm}
    \item We propose a novel ILA module to address the attention collapse problem faced by frozen ViTs in UFGIR tasks. The proposed ILA employs dual spatial-down sampling branches to aggregate discriminative features and reduce computational cost.
    \item We conduct comprehensive experiments across ten UFGIR datasets comparing more than 15 SotA methods across two image sizes. Through our experiments the proposed ILA obtains outstanding classification performance and enhanced computational efficiency, as measured by the total number of trainable parameters (TTP) and floating-point-operations (FLOPs). In particular, our method increases the average accuracy by at least 6.8\% compared to other methods in the parameter-efficient setting while requiring at least 123x less TTPs compared to current SotA UFGIR methods, and reducing the FLOPs by 30\% in average compared to other methods.
\end{enumerate}

\section{Related Work}
\label{sec_related}

\subsubsection{Ultra Fine-Grained Image Recognition}
\label{sssec_ufgir}

UFGIR methods employ backbones pretrained for generic recognition and equip them with modules to select and aggregate discriminative features \cite{he_transfg_2022,wang_feature_2021} or employ loss functions and tasks \cite{yu_maskcov_2021,yu_spare_2022} to guide models to more effectively make use of fine-grained features, or both \cite{wei_fine-grained_2021,fang_learning_2024}. In the former category, FFVT \cite{wang_feature_2021} employs ViT's attention scores to select intermediate low-, medium-, and high-level features that are vital for recognizing small inter-class differences and are aggregated through the last transformer encoder block.

Since UFGIR has an additional challenge due to limited labeled data, research in this direction has been very active in recent years \cite{yu_patchy_2020,yu_benchmark_2021}. MaskCOV \cite{yu_maskcov_2021}, SPARE \cite{yu_spare_2022}, and Mix-ViT \cite{yu_mix-vit_2023} employ data augmentation and propose self-supervised \cite{balestriero_cookbook_2023} tasks and losses for the model to learn intrinsic details with limited data. CLE-ViT \cite{yu_cle-vit_2023} and CSD \cite{fang_learning_2024} employ contrastive learning \cite{chen_simple_2020} and the latter also employs self-knowledge distillation \cite{hinton_distilling_2015,zhang_self-distillation_2022,caron_emerging_2021} to address the challenges faced in UFGIR. However, most of these methods employ ViT backbones with large number of parameters that all need to be stored for deployment and also spend significant resources during training.

\subsubsection{Parameter-Efficient Transfer Learning}
\label{sssec_petl}

(PETL) techniques aim to fine-tune a small subset of modules while most of the backbone parameters are frozen. These are mostly classified into two: prompt-tuning and adapters. Prompt-tuning \cite{lester_power_2021,avidan_visual_2022} incorporates additional task-specific learnable tokens that are appended to the sequence at different stages of the transformer. VQT \cite{tu_visual_2023} proposes using the tokens as queries that aggregate layer-wise information and are expedited to the classification layer to incorporate intermediate features into the classification head. However, incorporating additional tokens increases the computational cost of the forward pass, and in the case of VQT, the integration of a large number of input features into the classification head can rapidly increase the number of parameters.

On the other side, adapters were first proposed by Houlsby \etal \cite{houlsby_parameter-efficient_2019} in the natural language processing (NLP) domain and are additional light-weight non-linear modules that are inserted usually inside transformer layers. ConvPass \cite{jie_convolutional_2022} extended this idea for ViTs by incorporating a 2D convolution into an adapter to introduce spatial biases into the design.

\section{Method}
\label{sec_method}

\begin{figure}[!htb]
    \vspace{-0.5cm}
    \begin{center}
        \includegraphics[width=0.7\linewidth]{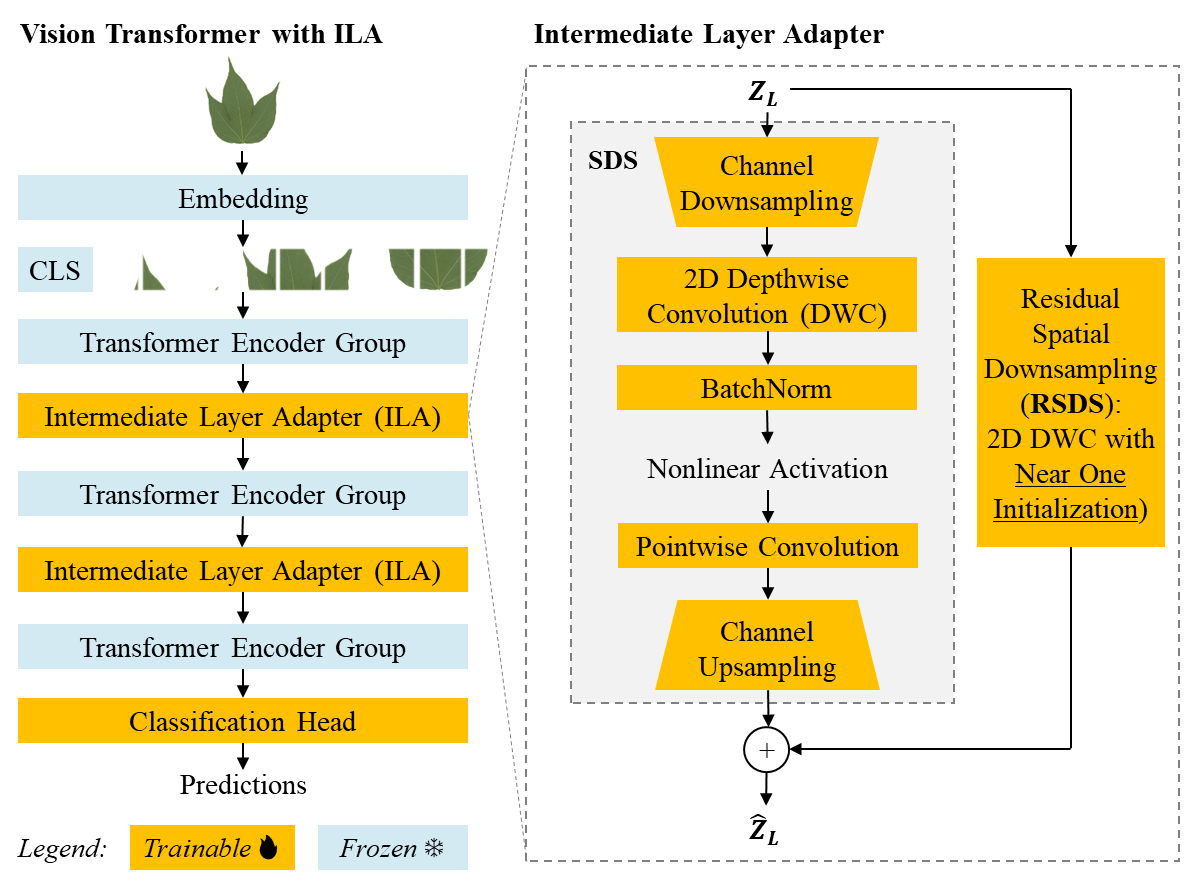}
    \end{center}
    \vspace{-0.5cm}
    \caption{Overview of ViT with our proposed Intermediate Layer Adapter (ILA). Trainable modules are shown in orange while frozen ones are shown in blue. An image is embedded into tokens and forwarded through a series of transformer encoder blocks, which we divide into three groups. After the first two encoder groups the sequence is passed through the ILA. After passing through all the encoder blocks the CLS token is forwarded through a classification head to obtain predictions. In the ILA tokens are forwarded through two spatial downsampling (SDS) branches. In the main SDS branch (highlighted as a grey box) tokens are first downsampled channel-wise and then spatially downsampled through the usage of a 2D depth-wise convolution.  The sequence is then forwarded through a BatchNorm layer, a non-linear activation, and a point-wise convolution, before being up-sampled channel-wise. To allow for residual gradient flow we also forward the tokens through a Residual Spatial Downsampling (RSDS) branch implemented as a 2D depth-wise convolution initialized with values near one. Initializing the kernel to values near one allows the RSDS to behave as a learnable identity or pooling function. Then, the outputs of the dual SDS branches are added together and forwarded to the next encoder group.}
    \vspace{-0.5cm}
    \label{figure_overview}
\end{figure}

The overview of our proposed method is shown in \Cref{figure_overview}. Our method is based on a generic Vision Transformer (ViT) \cite{dosovitskiy_image_2020}. An image is patchified and forwarded through transformer encoder blocks which we divide into three groups, each with 4 blocks. After the first two encoder groups the sequence is passed through the proposed Intermediate Layer Adapter (ILA). After passing through all the encoder blocks the CLS token is forwarded through a classification head to obtain predictions.

\subsubsection{Vision Transformer Encoder}
\label{sssec_vit}

Images are patchified using a convolution with kernel size $P$ and flattened into a 1D sequence of $D$ channels with length $N_0 = (h / P) \times (w / P)$, where $h$ and $w$ represent the image width and height. A learnable CLS token \cite{devlin_bert_2019} is appended to the sequence and learnable positional embeddings are added to encode spatial information. This sequence is passed through a series of transformer encoder blocks each composed of multi-head self-attention (MHSA) and position-wise feed-forward networks (PWFFN) \cite{vaswani_attention_2017}. The output of each block is $\mathbf{z}_l \in \mathbb{R}^{N_l\times D}$  Finally, this output is passed through a LayerNorm \cite{ba_layer_2016} and a linear classification layer to obtain predictions.

\begin{figure}[!htb]
    \vspace{-0.5cm}
    \centering
    \begin{minipage}{0.39\textwidth}
        \centering
        \includegraphics[width=\textwidth]{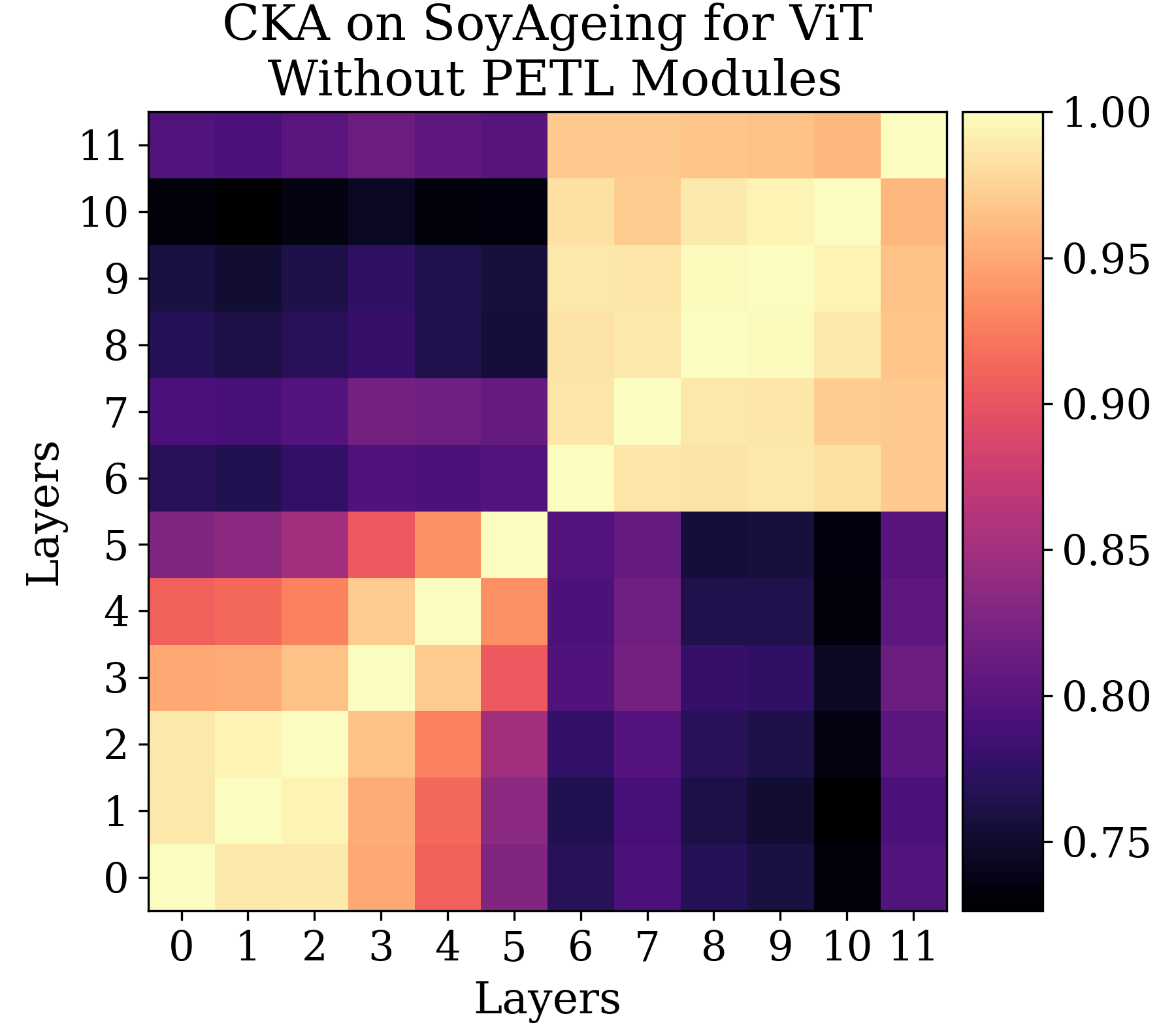}
        \label{figure_cka_attention_soyageing_vanilla}
    \end{minipage}
    \hfill
    \begin{minipage}{0.39\textwidth}
        \centering
        \includegraphics[width=\textwidth]{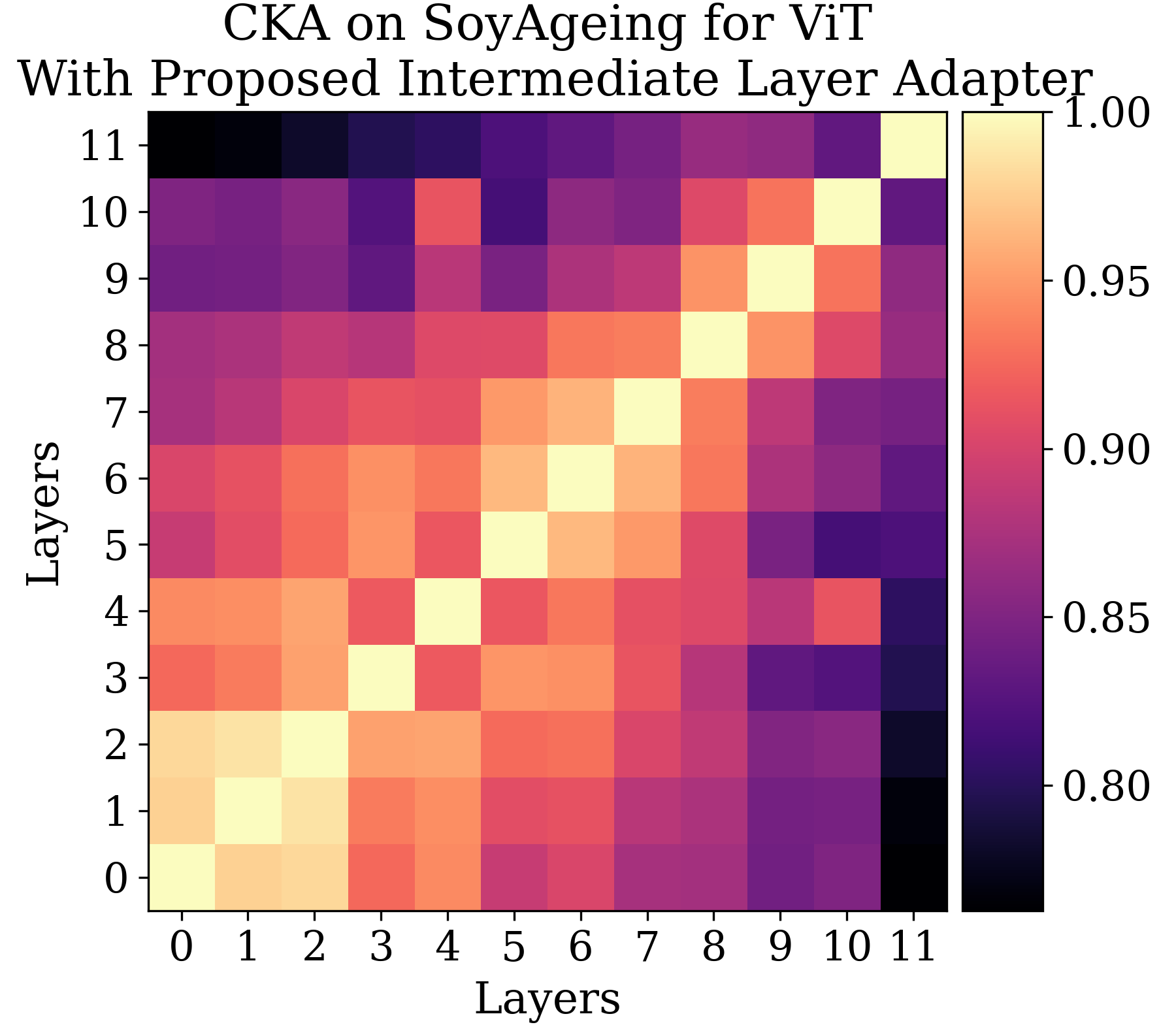}
        \label{figure_cka_attention_soyageing_ila}
    \end{minipage}
    \vspace{-0.5cm}
    \caption{Centered Kernel Alignment (CKA) similarity \cite{kornblith_similarity_2019} between attention layers of a ViT for the vanilla ViT (left) and ours (right). Lighter colors indicates higher similarity.}
    \vspace{-0.5cm}
    \label{figure_cka_attention}
\end{figure}

However, we observe that a frozen ViT encoder applied in UFGIR tasks suffers from  attention collapse \cite{zhou_deepvit_2021,wang_anti-oversmoothing_2022} as shown in \cref{figure_cka_attention}. This happens when attention maps across different layers collapse to a single representation and therefore the model is unable to extract meaningful features.

\vspace{-0.5cm}

\subsubsection{Inter-Layer Adapter}
\label{sssec_ila}

Inspired by previous works \cite{liu_swin_2021,wang_pyramid_2021} and to encourage the network to focus on distinct areas at different stages of encoding, we explicitly enforce hierarchy in the feature maps by incorporating downsampling in the ILA module. Specifically, we make use of a dual spatial down-sampling (SDS) design. The main branch is composed of a channel down-sampling (CDS), a depth-wise separable convolution where the spatial downsampling takes place, and a channel up-sampling (CUS) module. The design of this branch is similar to the one proposed by Jie \etal. \cite{jie_convolutional_2022}, but we have two important differences: 1) to increase computational efficiency we employ a depth-wise separable convolution \cite{howard_mobilenets_2017} for the convolution in main branch, 2) we do not incorporate padding therefore the forward through the convolution reduces the spatial dimension of the inputs. These design changes allow our model to not only reduce the cost, but also increase diversity in the attention maps by explicitly enforcing a hierarchy in the feature maps, which forces the model to focus on different areas at different stages. The output of this branch is denoted as $\mathbf{m}$ and defined as follows:

\begin{equation}
    \mathbf{m} = \text{CUS}(\text{PWConv}(\text{GELU}(\text{BN}(\text{DWConv}(\text{CDS}(\mathbf{z}_l))))))
    \label{eq_sds_main}
\end{equation}

\subsubsection{Residual Spatial Downsampling Branch}
\label{sssec_rsds}

To facilitate the smooth flow of information between layers and mitigate the risk of vanishing gradients within the network \cite{he_deep_2015} we wish to employ a residual connection. However, due to the spatial downsampling operation in \cref{eq_sds_main}, the spatial dimensions of $\mathbf{m}$ are reduced, preventing directly adding $\mathbf{z}_l$. To align their shapes, existing methods typically employ pooling or interpolation operations for downsampling $\mathbf{z}_l$. However, these apply fixed procedures that may discard local features or structural details. Also, they lack the learnability that has allowed neural networks to thrive in recent years \cite{krizhevsky_imagenet_2012,vaswani_attention_2017}.

Therefore, we propose employing a learnable residual downsampling branch based on depthwise convolutions (DWC) with kernel weights initialized close to 1, which can easily approximate an identity. We term this the Residual Spatial Downsampling (RSDS) branch. The equation for the residual output $r_{d,n}$ for each channel $d$ and each spatial position $n$ of a 1D depthwise convolution (DWC) with $D$ input and output channels and the input having $N$ spatial positions is shown in \cref{eq_dw_1d}:

\begin{equation}
    r_{d,n} = \sum_{k=0}^{K-1} z_l^{d,n+k} \cdot W_{d,k}; \qquad d=0,1,...,D; \qquad n=0,1,...,N
    \label{eq_dw_1d}
\end{equation}

In the case of kernel size $K=1$ and initializing the values of the kernel weights $W$ to values close to 1 the equation becomes close to an identity function $I$ as shown in \cref{eq_dw_approach_identity}:

\begin{equation}
        r_{d,n} \approx I(z_l^{d,n}) \approx z_l^{d,n}; \qquad d=0,1,...,D; \qquad n=0,1,...,N
    \label{eq_dw_approach_identity}
\end{equation}

However, unlike the identity function, the DWC can adapt its weights to regulate the influence of the original values in the overall output, effectively behaving as a gate \cite{hochreiter_long_1997}. This allows our module to learn different functions based on the gradients. For the case of kernel size different to 1, the proposed DWC with near ones init behaves similarly to a sum-pooling operation. Based on these observations and inspired by how we moved from fixed, manually-designed filters to learnable filters with AlexNet \cite{krizhevsky_imagenet_2012}, we employ this DWC with near ones initialization to act as a learnable residual.

\section{Experiment Methodology}
\label{sec_experiments}

Detailed description for our experiments can be found in the Appendix. We evaluate our method on ten ultra-fine-grained leaves datasets collected by Yu \etal \cite{yu_benchmark_2021} where each category represents a cultivar. When applicable, the best values are highlighted in \textbf{bold} and the second best are \underline{underlined}.

We report results using top-1 accuracy (\%) and standard deviation of 3 seeds for image size 224 and image size 448. We also report the number of trainable parameters (TTP) for a group of tasks and the number of floating-point operations (FLOPs). We use Pytorch \cite{paszke_pytorch_2019} and Wandb \cite{biewald_experiment_2020} to implement and manage experiments, respectively.

All of our experiments employ the ViT B-16 \cite{dosovitskiy_image_2020} backbone with patch size 16, number of layers $L=12$ and hidden dimension size $D=768$. We propose 3 different variants of ILA which are as follows:
\begin{itemize}
    \vspace{-0.3cm}
    \item ILA: the intermediate layer adapters (ILAs) modules with down-sampling are inserted after layer 4 and 8 only.
    \item ILA\textsuperscript{+}: includes the modules from ILA plus additionally ILAs without down-sampling are inserted at every other layers of the ViT model besides from layer 4 and 8. 
    \item ILA\textsuperscript{++}: includes the modules from ILA plus we additionally incorporate the traditional intra-layer adapters \cite{houlsby_parameter-efficient_2019} in all layers.
    \vspace{-0.3cm}
\end{itemize}
We compare our proposed models against 15 state-of-the-art models in the parameter-efficient setting, grouped into three families based on their characteristics: 1) methods which only fine-tune the classification head, 2) FGIR methods where a module is designed to explicitly select features based on some criteria, and 3) dedicated PETL methods.

\begin{table}[!htb]
    \centering
    \caption{Top-1 accuracy (\%) and total trainable parameters (TTP, in millions for all five tasks) for SotA models on five ultra-FGIR datasets with image size 448. Model* represents all parameters, including the backbone, were fine-tuned.}

    \begin{tabularx}{\linewidth}{X Z Z Z Z Z Z}
        \toprule
        Method & Cotton & SoyAgeing & SoyGene & SoyGlobal & SoyLocal & TTP ($10^6$) \\
        \midrule
        ViT B-16 & 39.03 & 48.61 & 21.31 & 24.97 & 28.72 & \textbf{1.7} \\
        MPNC \cite{li_towards_2018} & 43.89 & 40.43 & 20.22 & 25.79 & 27.94 & 4.9 \\
        IFA \cite{rios_anime_2022} & 44.58 & 56.01 & 33.09 & 35.82 & 29.22 & \textbf{1.7} \\
         \midrule
        TrFG \cite{he_transfg_2022} & 51.67 & 57.54 & 38.79 & 45.35 & 38.06 & 37.2 \\
        FFVT \cite{wang_feature_2021} & 51.94 & 69.93 & 49.97 & 47.70 & 42.22 & 37.2 \\
        CAL \cite{rao_counterfactual_2021} & 44.03 & 51.16 & 23.03 & 34.98 & 31.61 & 55.6 \\
        RAMS \cite{hu_rams-trans_2021} & 38.47 & 50.73 & 25.83 & 25.17 & 29.67 & \textbf{1.7} \\
        GLSim \cite{rios_global-local_2024} & 45.70 & 56.58 & 33.04 & 39.85 & 29.95 & 37.2 \\
         \midrule
        VQT \cite{tu_visual_2023} & 50.97 & 65.65 & 36.80 & 33.99 & 36.33 & 106.0 \\
        VPT-S \cite{avidan_visual_2022} & 38.19 & 49.12 & 27.43 & 27.37 & 28.39 & \underline{2.1} \\
        VPT-D \cite{avidan_visual_2022} & 43.19 & 60.76 & 38.28 & 36.83 & 25.28 & 6.3 \\
        ConvP \cite{jie_convolutional_2022} & 48.33 & 60.18 & 53.43 & 45.13 & 34.22 & 3.4 \\
        ADPT \cite{houlsby_parameter-efficient_2019} & 48.19 & 73.31 & 57.04 & 47.27 & 36.83 & 3.3 \\
         \midrule
        TrFG* \cite{he_transfg_2022} & 54.58 & 72.16 & 22.38 & 21.24 & 40.67 & 434.3 \\
        SIMT* \cite{sun_sim-trans_2022} & 54.58 & 34.76 & 15.46 & \textbf{70.69} & 25.00 & 497.0 \\
        CSD* \cite{fang_learning_2024} & \textbf{57.92} & \textbf{75.39} & \textbf{70.82} & 56.30 & 46.17 & 432.0 \\
         \midrule
        ILA\textsuperscript{+} & 53.33 & 68.79 & 52.65 & 48.29 & \underline{46.56} & 2.6 \\
        ILA\textsuperscript{++} & \underline{55.42} & \underline{75.00} & \underline{62.19} & \underline{58.14} & \textbf{50.83} & 3.5 \\
        \bottomrule
    \end{tabularx}

    \label{table_acc_params_cotton_soy}

\end{table}

\section{Results and Discussion}
\label{sec_results}

\subsubsection{Comparison with State-of-the-Art}
\label{sssec_discusion_sota}

A summary of aggregated results under the PE setting are shown in \cref{figure_accuracy_vs_flops_vs_params}. We observe that not only do the different versions of ILA achieve the top average accuracies across all tasks, but it is also parameter and compute efficient. Specifically, ILA\textsuperscript{++} increases accuracy by 6.8\% but requires 8\% less floating-point operations (FLOPs) and trains 90\% less parameters than FFVT \cite{wang_feature_2021}, which achieves the second highest average accuracy.

We also report per-dataset accuracies on \cref{table_acc_params_cotton_soy}, including fine-tuned (FT) FGIR models. While ILA does not obtain the best accuracy when compared to the best FT FGIR models it achieves a competitive accuracy at a much lower parameter-cost. Specifically, while the accuracy of ILA\textsuperscript{++} is 1\% lower compared to CSD \cite{fang_learning_2024} we remark that our model requires 123x less trainable parameters compared to CSD. Furthermore, as CSD proposes a self-supervised (SSL) and knowledge distillation (KD) enhanced training recipe, future work could aim to combined such SSL and KD recipes with ILA for improved performance.

\begin{table}[!htb]
    \centering
    \caption{Ablation on the design of the Residual Spatial Downsampling (RSDS) branch in terms of absolute top-1 accuracy and absolute difference with respect to the baseline.}
    \label{table_ablations}
    \begin{tabularx}{\linewidth}{X z z z z}
        \toprule
        \multirow{2}{*}{Model} & \multicolumn{2}{c}{SoyGlobal} & \multicolumn{2}{c}{SoyLocal}\\
        \cmidrule(lr){2-3} \cmidrule(lr){4-5}
         & Acc. (\%) & Diff. (\%) & Acc. (\%) & Diff. (\%) \\
         \midrule
         Baseline (ViT B-16) & $17.88\pm{0.40}$ &  & $28.83\pm{1.04}$ &  \\
        \midrule
        No RSDS & $2.16\pm{0.89}$ & -15.72 & $6.56\pm{1.42}$ & -22.27 \\
        \midrule
        RSDS: AvgPool & $29.07\pm{0.38}$ & 11.19 & $31.22\pm{1.25}$ & 2.39 \\
        RSDS: Convolution & $34.93\pm{1.37}$ & 17.05 & $27.83\pm{5.63}$ & -1.00 \\
        RSDS: DWC (Normal Init) & $29.16\pm{1.78}$ & 11.28 & $25.22\pm{1.95}$ & 3.61 \\
        RSDS: DWC (Near Ones) & $\mathbf{43.48\pm{0.21}}$ & \textbf{25.60} & $\mathbf{41.28\pm{0.98}}$ & \textbf{12.45} \\
        \bottomrule
    \end{tabularx}
    \vspace{-0.5cm}
\end{table}

\subsubsection{Ablation on Design of RSDS}
\label{sssec_ablations}

Results are shown in \cref{table_ablations}. We compare the baseline against a model where the skip connection is forfeited (No RSDS), and four variations of the RSDS module: based on average pooling, traditional convolution, depth-wise convolution with normal init, and our proposed depth-wise with near ones initialization. It is evident that the usage of RSDS is necessary to avoid collapse of the network. Furthermore, it is also evident that the proposed approach is more effective as a residual compared to others.

\section{Conclusion}
\label{sec_conclusion}

In this paper we propose a novel intermediate layer adapter based on dual-branch spatial down-sampling for parameter and compute efficient ultra fine-grained image recognition. The proposed approach increases the diversity in attention maps and obtains outstanding results in terms of accuracy-cost.

\section*{Acknowledgements}

This work was supported by the National Science and Technology Council, Taiwan, under grant NSTC 111-2221-E-A49-092-MY3, NSTC 113-2640-E-A49-005, NSTC 112-2221-E-007-079-MY3 and NSTC 113-2218-E-007-020. We also thank the National Center for High-performance Computing (NCHC) and NYCU HPC for providing computational and storage resources.

%
%
\bibliographystyle{splncs04}
\bibliography{main}

\clearpage
\appendix

\section*{Appendix for Down-Sampling Inter-Layer Adapter for \\ Parameter and Computation Efficient \\ Ultra-Fine-Grained Image Recognition}

\section{Experiment Methodology}
\label{sec_experiments}

In \cref{table_datasets} we describe the datasets used for our experiments. These are ultra-fine-grained leaves datasets collected by Yu \etal \cite{yu_benchmark_2021} where each category represents a confirmed cultivar name attached to the seed obtained from the genetic resource bank.

\begin{table}[!htb]
    \centering
    \caption{Dataset Statistics}
    \begin{tabularx}{\linewidth}{X Z Z Z}
        \toprule
        Datasets & Classes & Train Images & Test Images \\
        \midrule
        Cotton & 80 & 240 & 240 \\
        SoyAgeing & 198 & 4950 & 4950 \\
        SoyAgeingR1 & 198 & 990 & 990 \\
        SoyAgeingR3 & 198 & 990 & 990 \\
        SoyAgeingR4 & 198 & 990 & 990 \\
        SoyAgeingR5 & 198 & 990 & 990 \\
        SoyAgeingR6 & 198 & 990 & 990 \\
        SoyGene & 1110 & 12763 & 11143 \\
        SoyGlobal & 1938 & 5814 & 5814 \\
        SoyLocal & 200 & 600 & 600 \\
        \bottomrule
    \label{table_datasets}
    \end{tabularx}
\end{table}

We conduct our experiments in two stages: first we conduct a learning rate search, $LR \in (0.3, 0.1, 0.03, 0.01, 0.003)$  based on subsets from the train data to select the best learning rate. We select the $LR$ with highest accuracy in the validation subset. Then, in the second stage we use the LR from the first stage to train the model on the full training data and evaluate on the test set with 3 seeds. We use the Stochastic Gradient Descent (SGD) optimizer with momentum 0.9, batch size 8, cosine learning scheduler with 500 steps warm-up and we train all models for 50 epochs with automatic mixed-precision.

For data preprocessing, we resize our images to a square of size $300 \times 300$ or $600 \times 600$ and then crop a random square during training (or a center crop during inference) of size $224 \times 224$ or $448 \times 448$. All images are horizontally flipped and normalized based on standard ImageNet \textit{mean} and \textit{std} values.

We report results using top-1 accuracy (percentage) and standard deviation of the 3 seeds for image size 224 and image size 448. We also report computational cost based on a server with an RTX 3090 GPU. We report the number of trainable parameters (TTP) for a group of tasks, and the number of floating-point operations (FLOPs). We use Pytorch \cite{paszke_pytorch_2019} to implement our experiments and Wandb \cite{biewald_experiment_2020} for experiment managing.

When applicable, the best values are highlighted in \textbf{bold} and the second best are \underline{underlined}.

All of our experiments employ the ViT B-16 \cite{dosovitskiy_image_2020} backbone with patch size 16, number of layers $L=12$ and hidden dimension size $D=768$. We propose 3 different variants of ILA which are as follows:

\begin{itemize}
    \item ILA: the intermediate layer adapters (ILAs) modules with down-sampling are inserted after layer 4 and 8 only.
    \item ILA\textsuperscript{+}: includes the modules from ILA plus additionally ILAs without down-sampling are inserted at every other layers of the ViT model besides from layer 4 and 8. 
    \item ILA\textsuperscript{++}: includes the modules from ILA plus we additionally incorporate the traditional intra-layer adapters \cite{houlsby_parameter-efficient_2019} in all layers.
\end{itemize}

We compare our proposed models against 15 state-of-the-art models in the parameter-efficient, grouped into three families based on their characteristics. The first includes methods which only fine-tune the classification head. This includes the following:

\begin{itemize}
    \item Linear Classifier (Baseline): the most simple PE method which keeps all backbone parameters frozen and only fine-tunes the classification head.
    \item Low-Rank Bilinear Pooling (LR-BLP) \cite{kong_low-rank_2017}: employs a low-rank projection to reduce the dimensionality of the bilinearly pooled features.
    \item Matrix Power Normalized Covariance (MPN-Cov) \cite{li_is_2017,li_towards_2018}: applies covariance pooling of high-level features to lessen instabilities of bilinear pooling.
    \item Intermediate Features Aggregation (IFA) \cite{rios_anime_2022}: selects the CLS tokens from intermediate layers and forwards them through a small MLP to first aggregate cross-layer features before outputting classification predictions.
\end{itemize}

The second category includes FGIR methods where a module is designed to explicitly select features based on some criteria, along with a possible module to aggregate these selected discriminative features. We evaluate these models in the parameter-efficient setting (PEFGIR) where only a small percentage of modules are fine-tuned. It is composed of:

\begin{itemize}
    \item CAL \cite{rao_counterfactual_2021}: employs counter-factuality to train a bilinear attention module which is used for both pooling features and to generate augmented versions of the input images (crops and masked). The fine-tuned components include the attention module and the bilinear attention pooling classification head.
    \item TransFG \cite{he_transfg_2022}: selects features from the previous-to-last layer based on head-wise attention rollout \cite{abnar_quantifying_2020}, a matrix-multiplication based aggregation of attention scores across layers. We fine-tune the last transformer encoder block where the feature aggregation happens and the linear classification head.
    \item FFVT \cite{wang_feature_2021}: selects and aggregates intermediate features based on layer-wise attention. Same as the previous one: the last transformer encoder block and the classification head.
    \item RAMS-T \cite{hu_rams-trans_2021}: crops the image for data augmentation based on attention rollout \cite{hu_rams-trans_2021}. Fine-tunes only the classification head.
    \item GLSim \cite{rios_global-local_2024}: computes the similarity between global and local representations of an image to select crops. Fine-tunes an aggregator transformer encoder block and the classification head.
\end{itemize}

The third category includes dedicated PETL methods as follows:

\begin{itemize}
    \item VPT-Shallow (VPT-Sh) \cite{avidan_visual_2022}: appends learnable prompts to the sequence at the start of the transformer.
    \item VPT-Deep \cite{avidan_visual_2022}: appends learnable prompts to the sequence before each transformer block and then removes them after each block.
    \item Visual Query Tuning (VQT) \cite{tu_visual_2023}: appends learnable prompts to be used as queries only (not keys or values) to the sequence prior to the MHSA module of each layer. These prompts are expedited towards the classification head where they are concatenated into a single large dimensional linear layer.
    \item Adapter \cite{houlsby_parameter-efficient_2019}: incorporates a small MLP inserted after the MHSA and PWFFN of each transformer encoder block.
    \item ConvPass \cite{jie_convolutional_2022}: similar to the previous it incorporates a small MLP inserted inside the transformer block, but this MLP incorporates a $3\times 3$ convolution in between the channel downsampling and upsampling of the adapter. 
\end{itemize}

We integrate all the previous methods into our experiment framework to ensure consistent training and evaluation for a fair comparison. Asides from these, we also compare against results previously published in the UFGIR literature, in particular the results published by Fang \etal \cite{fang_learning_2024} which include:

\begin{itemize}
    \item SIM-Trans \cite{sun_sim-trans_2022}: incorporates structure information and multi-level feature contrastive learning.
    \item CSDNet \cite{fang_learning_2024}: incorporates contrastive learning and self-distillation for learning with limited samples.
\end{itemize}

\end{document}